\documentclass[letterpaper, 10 pt, conference]{ieeeconf}  
\usepackage{todonotes}
\usepackage{mathtools}
\usepackage{url}
\usepackage{siunitx}
\usepackage{amsmath,amssymb,relsize}
\newcommand{\allfor}{\displaystyle\mathop{\mathlarger{\forall}}}

\IEEEoverridecommandlockouts                              

\title{\LARGE \bf
A Self-Supervised Learning Approach to Rapid Path Planning  \\ for Car-Like Vehicles Maneuvering in Urban Environment
}

\author{Piotr Kicki$^{1}$, Tomasz Gawron$^{2}$ and Piotr Skrzypczyński$^{1}$
\thanks{$^{1}$ Institute of Robotics and Machine Intelligence; 
$^{2}$ Institute of Automation and Robotics, Poznan University of Technology, Poznan, Poland; {e-mail: \tt\small \{name.surname\}@put.poznan.pl}}%
}

\begin{document}

\maketitle
\thispagestyle{empty}
\pagestyle{empty}

\begin{abstract}

An efficient path planner for autonomous car-like vehicles should handle the
strong kinematic constraints, particularly in confined spaces commonly encountered while maneuvering in city traffic, and should enable rapid planning, as the city traffic scenarios are highly dynamic.
State-of-the-art planning algorithms handle such difficult cases at high computational cost,
often yielding non-deterministic results.
However, feasible local paths can be quickly generated leveraging the past planning
experience gained in the same or similar environment.
While learning through supervised training is problematic for real traffic scenarios,
we introduce in this paper a novel neural network-based method for path planning,
which employs a gradient-based self-supervised learning algorithm to predict feasible paths.
This approach strongly exploits the experience gained in the past and rapidly yields feasible maneuver plans for car-like vehicles with limited steering-angle.
The effectiveness of such an approach has been confirmed by computational experiments.
\end{abstract}

\section{Introduction}
Although path and motion planning is one of the most researched areas in robotics \cite{lavallebook},
new practical challenges, such as planning the motion of self-driving vehicles in city traffic
push the research into directions that are still unexplored.
Motion planning for self-driving cars requires fast planning under changing traffic
conditions, taking into consideration the constraints imposed by the vehicle and environment.
Classic motion planning methods, like sampling-based planning, are considered sufficient
in typical scenarios for car-like vehicles \cite{motionreview}, but their ability to handle highly constrained planning cases often comes at the cost of non-deterministic solutions and high computation cost that grows with the complication of the environment.
In contrary, human drivers efficiently and quickly plan the paths of their cars in short time horizon
even for highly constrained cases, which can be attributed to the use of prior experience on how to perform planning.
State-of-the-art path planning methods for car-like vehicles do not integrate the prior experience,
which may lead to failure in situations that require rapid response and reacting to the dynamically changing environment.
Considering this motivation we investigate how the experience gathered by a robotic vehicle while navigating in a given environment can be used to improve the efficiency of path planning in similar environments through the application of machine learning.

\begin{figure}[th]
    \centering
    \includegraphics[width=\linewidth]{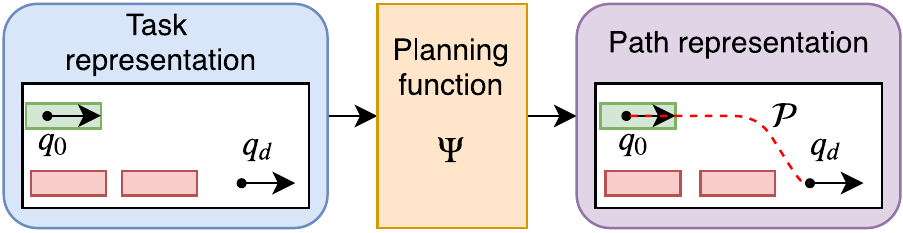}
    \caption{General scheme of the proposed approach to efficient path planning with the use of neural network trained in the self-supervised manner. Planning function $\Psi$, which transforms the representation of the task into representation of the path is approximated by the neural network.}
    \label{fig:scheme}
    \vspace{-0.3cm}
\end{figure}

We propose a neural network, which based on the representation of a task generates the representation of a feasible path (Fig. \ref{fig:scheme}).
To avoid laborious gathering of data from human drivers, which is a typical case in supervised end-to-end control solutions for self-driving cars \cite{deepdriving}, we exploit a self-supervised approach.

The main difficulties we have identified in implementing this idea are an efficient way to represent the environment (local map) at the input of the planning module,
and construction of the loss function that penalizes violation of the numerous constraints that are present in our scenario.
The former difficulty was overcome by applying to this practical problem our recently introduced general neural network architecture \cite{Ginv} that is invariant to the actions of symmetric subgroups, thus being able to represent correctly the shapes of obstacles provided as vectors of 2-D coordinates.
The later problem is solved by using a differentiable loss function that explicitly takes into account the geometry of the planned path and penalizes violation of the kinematic constraints and the collisions.
These components, together with the idea of self-supervised planning under differential constraints by approximating the \textit{oracle planning function}, are considered the main contribution of the proposed planning method, which is studied in this paper through a series of computational experiments.
We demonstrate the generalization abilities of our network, illustrate its advantages over selected state-of-the-art planning algorithms, and quantitatively analyze the accuracy of the proposed solution, planning times and lengths of the generated paths.

\section{Related Work}
Much of prior work on path planning addresses algorithms for discrete state spaces
that search on graphs and scale poorly with the increasing dimensionality \cite{lavallebook}.
Although the classic A$^*$ search algorithm was successfully adopted to the kinematic state
space of a car and used as part of the motion planner by the DARPA Urban Challenge winning entry \cite{thrun}, sampling-based algorithms are more computationally efficient in high dimensional spaces or
when confronted with complex environments \cite{karaman}.
Sampling-based algorithms most widely used in robotics belong to the Rapidly-exploring Random Trees (RRT) \cite{lavallebook} family.
They have many variants, such as RRT$^*$ \cite{karaman} that yields the optimal (shortest) path
with asymptotical guarantees, or RRT-path \cite{vonasek2009} that improves the algorithm efficiency in narrow passages.
While both optimality of the generated path and good performance in confined spaces are of high importance for motion planning in the urban environment, the performance of RRT-like sampling-based planners suffers when nonholonomic constraints and steering angle limits are present in the vehicle \cite{NonholonomicRRT}.
Moreover, none of the classic methods, either deterministic or randomized can learn
from past experience, which often leads to the time-consuming manual tuning of parameters in these methods in order to get them working with the particular vehicle model and environment.

The success of machine learning methods, and deep neural networks in particular, for applications in robotics \cite{dlimits} increased also the interest in robot motion planning methods involving machine learning or entirely learned from data.
Information from past searches in similar environments was used in \cite{zhang} to learn a sampling policy that accepts or rejects uniformly generated samples in a randomized path search algorithm, reducing computation time.
Berenson et al. \cite{berenson} exploited past experience in a way more similar to our ideas.
They stored entire paths and re-used or repaired the existing ones to reduce the planning time in high-dimensional spaces of manipulation tasks.
Recurrent neural networks were employed in the OracleNet \cite{oracle2019} to generate new paths
using paths obtained from demonstrations.
While this approach was an inspiration for our work, it differs significantly
from the planner presented in this paper, because it does not use any environment model,
relying on the demonstrated paths to explore the state space.
Such an approach is rather infeasible for vehicles in urban scenarios, where the
state space may differ considerably for particular maneuvers and local scenes.
Therefore, while we retain the notion of ``oracle'', we use an entirely different
neural network architecture and the self-supervised learning schema instead of demonstrations.
The OracleNet concept was then extended to the Motion Planning Networks (MPNet) framework \cite{mpnet},
which generates collision-free paths for the given start and goal states (configurations)
directly taking point clouds as the input.
Unlike OracleNet, MPNet generalizes to unseen environments, but still learns through imitation,
which imposes high demands on the data collection procedure in the case of autonomous vehicles.
This problem is negligible in our approach due to self-supervised learning using simple environment models.

Recently, a number of learning-based methods specialized in path planning for mobile robots and autonomous vehicles have emerged.
Point-to-point and path-following navigation behaviors are learned with AutoRL \cite{autorl}.
This approach leverages reinforcement learning and adds an evolutionary automation layer
to find a reward and then a neural network architecture (hyper-parameters) that maximizes the cumulative reward.
Although the learned motion behaviors are smooth and safe, they are expensive to train, as the method requires a large amount of input data due to the size of learnable parameters.
A jointly learnable behavior and trajectory planner for self-driving vehicles was proposed in \cite{urtasun}.
This approach learns a shared cost function used to select behaviors that handle traffic rules
and generate vehicle trajectories employing a continuous optimization solver.
Nevertheless, it learns from a database of human driving recordings rather than from its own past experience.

\section{Problem Statement}
\subsection{Planning Problem Formulation}
\label{sec:problem_formulation}
The problem considered in this paper is to plan a feasible monotonic path from an initial state $q_0$ to the desired state $q_d$ in some environment $E$, taking into account the vehicle kinematics, physical dimensions and its limited steering angle $\beta$.

\subsubsection{Model of the Vehicle}
The vehicle considered in this paper is a typical car. First, we define its state by
\begin{equation}
    q = \begin{bmatrix}\beta & \theta & x & y\end{bmatrix}^T,
\end{equation}
where $\beta$ is a steering angle of the virtual steering wheel, $\theta$ is an orientation of the vehicle, $x$ and $y$ are the coordinates of the guiding point $P$ in the world coordinate system. The model of the car kinematics is described by
\begin{equation}
    \dot{q} = \begin{bmatrix}\dot{\beta}\\ \dot{\theta}\\ \dot{x}\\ \dot{y}\end{bmatrix} = 
    \begin{bmatrix}
        1 & 0 \\
        0 & \frac{1}{L}\tan\theta\\
        0 & \cos\theta\\
        0 & \sin\theta\\
    \end{bmatrix} \begin{bmatrix}\zeta\\ v\end{bmatrix},
\end{equation}
where $L$ is a distance between front and rear wheels, $\zeta$ is an angular speed of the steering wheel and $v$ is a longitudinal velocity of the vehicle guidance point.
We assume no longitudinal and transverse slip of the wheels and neglect the dynamics of the car as well. However, we model the constraint put on the steering wheel angle $\beta$, which is limited by the maximum steering wheel angle $\beta_{max} > 0$, such that $|\beta| < \beta_{max}$.

\begin{figure}[th]
    \centering
    \includegraphics[width=0.8\linewidth]{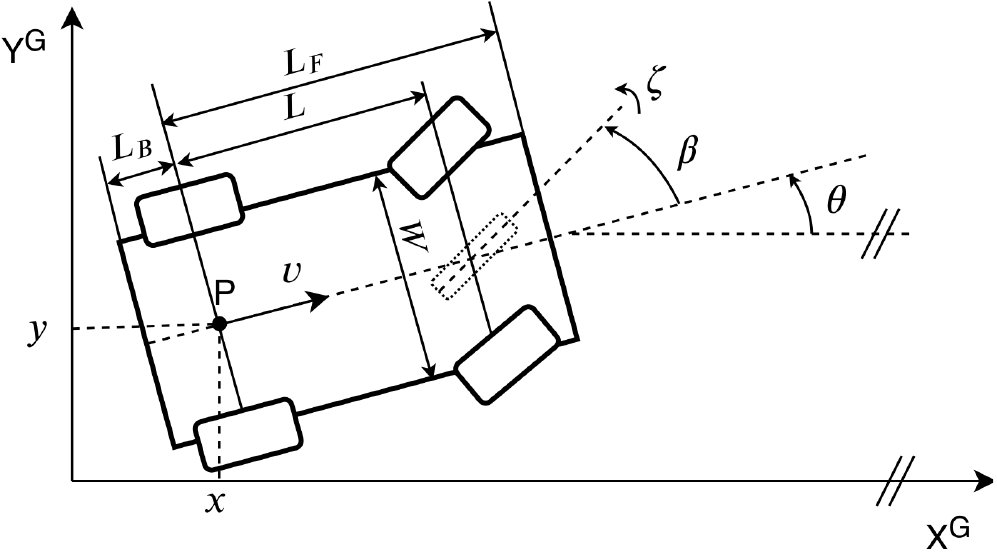}
    \caption{Model of the considered vehicle.}
    \label{fig:car}
    \vspace{-0.3cm}
\end{figure}

We approximate the typical body of a car with a rectangle with width $W$ and length $L_B + L_F$. All the aforementioned physical dimensions, state coordinates as well as model inputs are depicted in Figure \ref{fig:car}.

\subsubsection{Model of the Environment}
We model environment $E$ of the vehicle by the free space $\mathcal{F}$, which is a logical sum of at most $N_q$ quadrangles
\begin{equation}
    E = \mathcal{F} = \bigcup_{i=1}^{N_q} \mathcal{F}_i,
\end{equation}
where $\mathcal{F}_i$ denotes the $i$-th quadrangle. 
This kind of representation allows to model typical scenarios in man-made environments while keeping the local environment map very simple.
It is worth to notice that the proposed environment model can be easily produced from either visual or range data, and is not specific to any sensing modality. 

\subsubsection{Path Representation}
In this paper, paths are represented as a spline with $N$ segments defined as 5-th degree polynomials, as they may be tracked by a car with a continuous steering angle.
Such a spline curve can be described as a concatenation of $N$ polynomials of 5-th degree, taking the form:
\begin{equation}
    \mathcal{P}_i(t) = (x(t), y(t))\quad \text{for}\quad i=1,2,\ldots,N,
\end{equation}
where $x(t)$ and $y(t)$ are some parametric 5-th degree polynomials parameterized with $t \in [0;1]$, which satisfy the following relationship
\begin{equation}
    \allfor_{i=1,2,\ldots,N-1} \allfor_{k=0,1,2} \mathcal{P}_i^k(1) = \mathcal{P}_{i+1}^k(0),
\end{equation}
where $\mathcal{P}_i^k$ denotes the $k$-th derivative of $\mathcal{P}_i$ with respect to the parameter $t$. Those points, where two consecutive polynomials $\mathcal{P}_i$ and $\mathcal{P}_{i+1}$ meet each other are named \textit{gluing points}.

\begin{figure}[th]
    \centering
    \includegraphics[width=\linewidth]{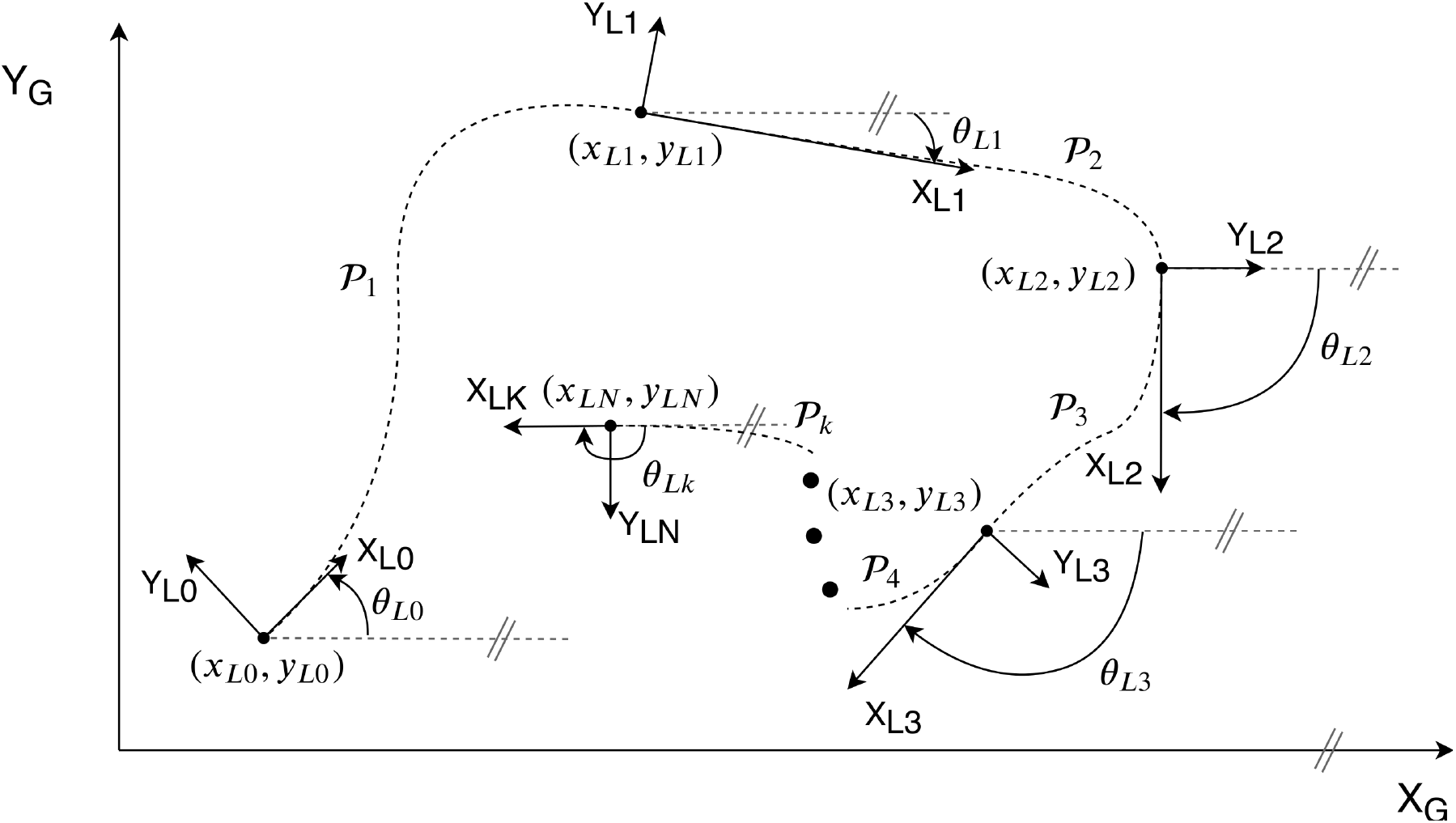}
    \caption{Visualization of the proposed representation of the spline built with 5-th degree polynomials defined in the local coordinate systems associated with the gluing points.}
    \label{fig:lcs}
    \vspace{-0.4cm}
\end{figure}

To form a valid path in the considered scenario, the spline $\mathcal{P}$ has to meet the boundary constraints defined by the initial and goal states.
Thus, to fully define a path $\mathcal{P}$ in a context of specific task definition we have to determine $N-1$ gluing points, as the initial and goal points are defined in the task. Therefore, the path $\mathcal{P}$ can be defined as a vector of $N-1$ gluing points
\begin{equation}
\label{eq:path_definition}
    \mathcal{P} = [p_1\, p_2\, \ldots\, p_{N-1}],
\end{equation}
where each gluing point $p_i$ is given by
\begin{equation}
p_i = \left(x_{Li}, y_{Li}, \frac{dy_{Li}}{dx_{Li}},\frac{d^2y_{Li}}{dx_{Li}^2} \right),
\end{equation}
where $x_{Li}$ and $y_{Li}$ denote the $x$ and $y$ coordinates of the $i$-th gluing point, while $\frac{dy_{Li}}{dx_{Li}}$ and $\frac{d^2y_{Li}}{dx_{Li}^2}$ denote the first and second derivative of the path in the $i$-th gluing point, respectively. 
All values describing $i$-th gluing point are expressed in the local coordinate system associated with $(i-1)$-th gluing point in order to enable the path $\mathcal{P}$ to be represented in the control-law as the level-curve \cite{levelcurve}. As a result, segments are described by the 5-th degree polynomials in the local coordinate system associated with the $(i-1)$-th gluing point, defined as
\begin{equation}
\label{eq:P_i}
    \mathcal{P}_i(x) = \sum_{j=0}^{5} a_{ij} x^j,
\end{equation}
where $a_{ij}$ are coefficients of $i$-th polynomial.
Visualization of an example spline $\mathcal{P}$ constructed from the polynomials $\mathcal{P}_i$ in the local coordinates systems associated with the gluing points is shown in Fig. \ref{fig:lcs}.

\subsection{Dataset}
\label{sec:ds}

We considered 3 general planning scenario types:  overtaking maneuvers, perpendicular parking and oblique parking. When a planning scenario is generated, the type is chosen and the dimensions of initially chosen quadrangles are randomly modified (up to $\SI{10}{\metre}$ size differences and $90^\circ$ parking angle variations). After that, we sample initial and final vehicle states and check (using grid search) if a feasible Dubins path can be planned with a resolution of $\SI{0.4}{\metre}$.  We assumed zero initial/final steering angles since they do not affect vehicle body configuration.


The gathered data is split into 3 disjoint datasets: training, validation, and test. The training and validation dataset contains mostly the same environments (local maps), but different initial and final states. However, in the validation dataset, some environments were not present in the training dataset to keep track of the generalization to unseen environments, as the test set contains environments that are not present in both training and validation sets.
In the training set, there are 15432 pairs of initial and final states located in 208 environments, whereas in the validation dataset there are 4227 pairs of states in 214 environments. The test set consists of 1190 pairs of initial and final states, which are located in 60 randomly generated environments.

We evaluated path length and time elapsed to the first solution for our implementations of RRT* and State Lattice path planners. RRT* was endowed with Dubins distance metric and an extend procedure utilizing 5-th degree polynomials, for which the steering angles constraints were checked on the fly. State Lattice planner was carrying out an A* search with motion primitives consisting of 9 precomputed splines. Note that due to limited (finite) resolution of State Lattices and prohibitive computational cost of RRT* planner those planners have an accuracy less than 1, as they exceed the planning time limit (\SI{60}{\second}).

\section{Proposed Method}

In this section, we introduce a novel approach to path planning for robots with kinematic constraints, which exploits the neural network trained in a semi-supervised manner to generate a spline path for a given environment and both initial and desired final robot states.

\subsection{Idea of Path Planning with the Use of Neural Networks}

Let us consider the path planning function
\begin{equation}
    q_r = \Psi(q_0, q_d, E, V, \theta, \phi), \label{gendef}
\end{equation}
which for a given initial state $q_0$, desired state $q_d$, environment representation $E$, vehicle model $V$, some parameters $\theta$ and some random variable $\phi$ (taking $\phi$ as an argument allows incorporating the stochastic planners in the above mentioned definition) returns the path $q_r$. 
Such a planning function is a generalization of both stochastic and deterministic motion planning algorithms, as it can describe their operation, taking into account only the response of a given algorithm to the task of planning the path, and disregarding the internal structure of the given algorithm. Because the definition given in (\ref{gendef}) is very general, in this paper we limit ourselves to the case where the planning function is deterministic and the vehicle model $V$ is given, as it was stated in the section \ref{sec:problem_formulation}. Thus, our planning function simplifies to the form
\begin{equation}
    q_r = \Psi(q_0, q_d, E, \theta).
\end{equation}

Let consider a special planning function, which for every input always returns a feasible path if such a path exists, but does not guarantee
that the found path is optimal, e.g. with respect to its length. A function with this property we call \textit{oracle planning function}, as knowing the exact formula of such a function enables to solve all planning problems. However, it is difficult to give a closed formula for such an oracle function.

Despite this, human drivers can intuitively tell the rough shape of the path that is feasible for the tasks they repeat many times, e.g. while executing parking maneuvers.
This suggests that the oracle planning function may be approximated in some narrow range of its parameters, namely for short paths in some set of environments of fixed topology, using the prior experience.

Neural networks are the class of learnable models that are known for their ability to approximate nonlinear, complicated functions of many variables, once enough training examples are available. Moreover, their inference time is usually short and stable between calls.
Therefore, we propose to learn a deep neural model and use it as the oracle planning function estimate in the rapid path generation problem.

\subsection{Neural Network Architecture}
\label{sec:arch}
\begin{figure}[t]
    \centering
    \includegraphics[width=0.85\linewidth]{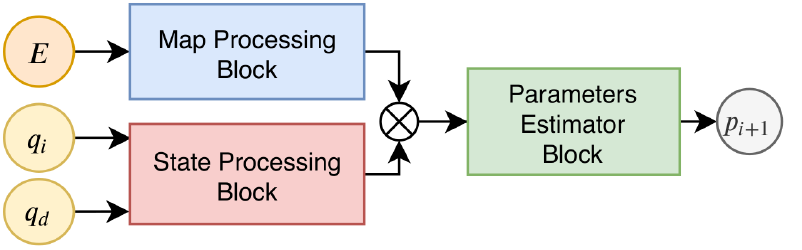}
    \caption{General scheme of the proposed path planning neural network architecture. Map Processing Block produce the latent representation of the environment $E$, whereas State Processing Block produces a latent representation of the actual $q_i$ and desired $q_d$ robot state. Then, both latent vectors are concatenated and the parameters $p_{i+1}$ of the $i+1$-th gluing point are determined with the use of Parameters Estimator Block.}
    \label{fig:arch}
    \vspace{-0.3cm}
\end{figure}

To approximate the oracle planning function $\Psi(q_0, q_d, E, \theta)$, we propose a neural network, which for a given set of arguments produce a representation of the path described in \eqref{eq:path_definition}. 
The assumed spline-based path representation affects the proposed neural network architecture. In fact, the proposed neural network is meant to estimate, at one time, the parameters of the next gluing point only. However, it is evaluated several times to obtain the next $n$ gluing points, which will lead the vehicle to the state (last gluing point), which can be directly connected to the desired state $q_d$ without violating the constraints with the use of the 5-th order polynomial. 
Proposed architecture is presented in Figure \ref{fig:arch} and consists of 3 functional blocks:
\begin{itemize}
    \item Map Processing Block, which given a map (defined as a set of convex quadrangles) produce its latent representation using the $G$-invariant network architecture \cite{Ginv} to obtain a latent representation of each quadrangle and the sum operation applied to the representations of quadrangles,
    \item State Processing Block, which given an actual $q_i$ and desired $q_d$ robot state produces a latent representation of the error,
    \item Parameters Estimator Block, which estimates the parameters $p_{i+1}$ of the $(i+1)$-th gluing point.
\end{itemize}
This network is used $n$ times to determine the next $n$ gluing points. Between network evaluations, a calculation of the next state $q_{i+1}$ (after the execution of the previously defined segment) is performed and that value is passed to the network input as the current state.

During the experiments, we use the following neural networks to implement functional blocks:
\begin{itemize}
    \item Map Processing Block - consists of the $G$-invariant neural network \cite{Ginv}, which input processing block consists of 2 fully connected (FC) layers with 32 and 128 neurons, and output block consists of 2 FC layers with 64 neurons each,
    \item State Processing Block - consists of 4 FC layers with 64 neurons in the first layer and 256 in each of the remaining layers,
    \item Parameters Estimator Block - consists of 4 neural networks, one for each parameter, each of them consists of 4 FC layers with 128, 64, 64 and 1 neuron in the subsequent layers.
\end{itemize}
All layers use $\tanh$ activation function, except the outputs of the Parameters Estimator Block which in general use an identity function in the last layer. Only untypical transformation is applied to the output of the estimator of the translation $x_i$ in the local coordinate system. To avoid negative values, as we assumed monotonic maneuvers, we apply sigmoid at the output and scale it 10 times. Moreover, we bias it by 0.1 to avoid the case when $x_i$ close to 0 leads to the badly conditioned matrix in the polynomials parameters determining procedure. 

\subsection{Loss Function}
\label{sec:loss}
In order to approximate the oracle planning function, the neural network needs a differentiable loss function
that penalizes infeasible paths. The generated paths are considered infeasible if they violate
constraints imposed either by the vehicle kinematics or the environment map.
The proposed loss function has 5 components: 
\begin{itemize}
    \item collision loss $\mathcal{L}_{coll}$, to force the planner to produce collision-free paths,
    \item curvature loss $\mathcal{L}_{curv}$, to ensure that the produced paths are possible to follow by the car with a limited steering angle,
    \item overshoot loss $\mathcal{L}_{over}$, to ensure that the last gluing point is possible to connect with the desired state $q_d$ without changing the sign of the velocity vector with the use 5-th degree polynomial,
    \item non-balanced loss $\mathcal{L}_{nbal}$, to ensure that the proposed segments have a similar length,
    \item length loss $\mathcal{L}_{len}$, to ensure that the feasible paths have also reasonable length,
\end{itemize}
which are summed together to obtain the overall loss 
\begin{equation}
    \mathcal{L} = \sigma_{coll}\mathcal{L}_{coll} + \mathcal{L}_{curv} + \mathcal{L}_{over} + \mathcal{L}_{nbal} + \sigma_{len}\mathcal{L}_{len},
\end{equation}
where $\sigma_{len}$ is a decision variable, which equals $1$ when $\mathcal{L}_{coll} + \mathcal{L}_{curv} + \mathcal{L}_{over} = 0$, what means that the returned path is feasible, and $0$ otherwise.
In turn, $\sigma_{coll}$ is a decision variable, which is set to $0$ in the pretraining phase and is set to $1$ in all other cases. the pretraining phase is applied at the very beginning to eliminate the most important loss $\mathcal{L}_{over}$ and to limit the segments length imbalance $\mathcal{L}_{nbal}$, and curvature $\mathcal{L}_{curv}$ temporarily without taking care of the obstacles.

\begin{figure}[ht]
    \centering
    \includegraphics[width=0.65\linewidth]{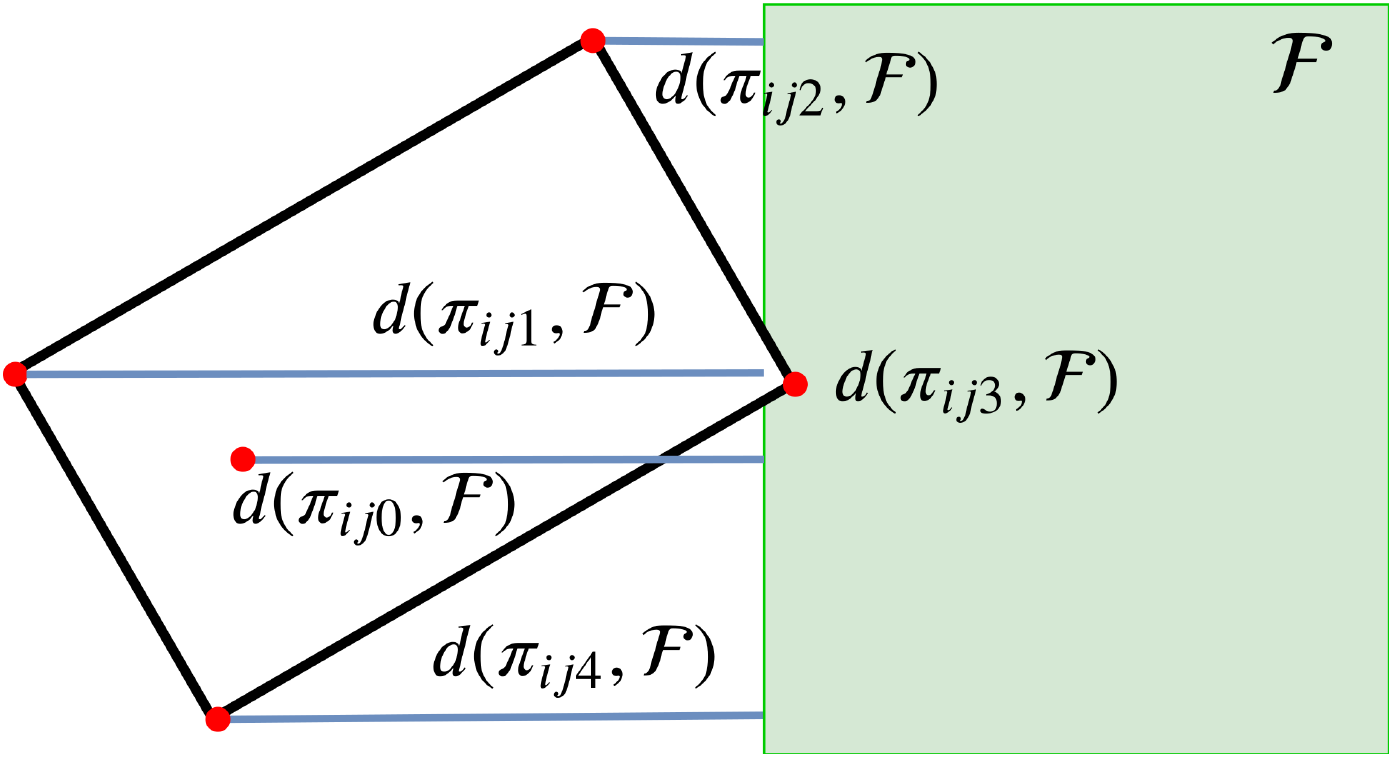}
    \caption{Visualization of the distances $d(\pi_{ijk}, \mathcal{F})$ (blue lines) from distinguished points $\pi_{ijk}$ (red dots) to the free space $\mathcal{F}$ (green area) used in the calculation of the collision loss $\mathcal{L}_{coll}$.}
    \label{fig:collision}
    \vspace{-0.2cm}
\end{figure}

The aforementioned losses are rather not standard in machine learning, so we describe bellow their definitions.
The collision loss $\mathcal{L}_{coll}$ is calculated by choosing 128 points on each of segments, such that the distance in the $X$ axis of the local coordinate system, in which that segment is defined, between adjacent points is constant. 
For each of those points, the 5 characteristic points $\pi_{ijk}$ on the vehicle model (four corners of the rectangular body of the vehicle and the guiding point in the middle of the rear axle) are calculated \cite{distpoints} and used for the collision loss calculation.
The resultant collision loss is defined by
\begin{equation}
    \sum_{i=1}^{N} \sum_{j=1}^{127} \sum_{k=1}^{5} d(\mathcal{F}, \pi_{ijk}) d(\pi_{ij0}, \pi_{(i+1)j0}),
\end{equation}
where $d(X, Y)$ denotes the smallest Euclidean distance between $X$ and $Y$. Index $i$ denotes the number of the segment, $j$ the number of point in the segment, whereas $k$ the number of the characteristic point ($0$ denotes the guiding point, points from $1$ to $4$ are the corners, numbered clockwise). An intuitive visualization of the loss calculation is depicted in Fig. \ref{fig:collision}.

The curvature loss $\mathcal{L}_{curv}$ is calculated by choosing 128 points on each of the segments (the same as for the collision loss) and then, knowing exact formula of the segment \eqref{eq:P_i}, the absolute value of its curvature is determined in all of those points $|\kappa_{ij}|$. 
The resultant curvature loss is defined by
\begin{equation}
    \sum_{i=1}^{N} \sum_{j=1}^{127} \max(|\kappa_{ij}| - \kappa_{max}, 0),
\end{equation}
where $\kappa_{max}$ is the maximal admissible curvature.

The overshoot loss $\mathcal{L}_{over}$ is calculated as the sum of three elements: (i) distance in the $X$ axis of the local coordinate system associated with $q_d$, from the last gluing point $p_{N-1}$ to the left half-plane of that local coordinate system, (ii) distance in the $X$ of the local coordinate system associated with $p_{N-1}$, from the desired point to the right half-plane of that local coordinate system, (iii) excess of difference in orientation between $p_{N-1}$ and $q_d$ over $\frac{\pi}{2}$.

The non-balanced loss $\mathcal{L}_{nbal}$ is given by
\begin{equation}
    \mathcal{L}_{nbal} = \sum_{i=1}^{N} \max(l_i - 1.5 \bar{l}, 0) + \max(\bar{l} - 1.5 l_i, 0),
\end{equation}
where $l_i$ denotes the length of the $i$-ith segment and $\bar{l}$ the mean length of all segments.

Finally, the length loss $\mathcal{L}_{len}$ is a simple sum of the lengths of the segments.

\section{Experiments}
To evaluate the proposed path planner we trained the neural network (with the architecture described in the Section \ref{sec:arch}) to estimate the 6 gluing points using the training set described in Section \ref{sec:ds}\footnote{The code of the implemented planner and the data used for training are available online at \url{https://github.com/Kicajowyfreestyle/tisaf}}. We used the curriculum learning technique \cite{curriculum} and train the network first to perform the overtaking maneuver, then perpendicular parking task, and finally angle parking. However, we conjecture that it is possible to train all types of maneuvers at once. The network was trained using Adam optimizer \cite{adam} with a learning rate equal $10^{-4}$ and batch size equal $64$. 
For the evaluation we choose the model with the highest accuracy on the validation dataset. In the presented experiments we use following values of the parameters introduced in Sections \ref{sec:problem_formulation} and \ref{sec:loss}: the number of segments $N = 7$, vehicle width $W = \SI{1.72}{\metre}$, distance from rear axle to the back $L_B = \SI{0.67}{\metre}$ and front $L_F = \SI{3.375}{\metre}$ and maximal admissible curvature $\kappa_{max} = \SI{0.22}{\per\metre}$.

We perform analysis of the proposed path planner by assessing its accuracy, planning time and the length of the returned paths in comparison with the popular path planning algorithms: State Lattices (SL) \cite{SL} and RRT* \cite{NonholonomicRRT}, which implementations are described shortly in Section \ref{sec:ds}. The results of that comparison are shown in Tab. \ref{tab:results}. The accuracy measure reported in Tab. \ref{tab:results} is the ratio of tasks for which the given method returned a feasible path, to the total number of tasks. As typical traffic scenarios require a fast response we set the limit to the planner running time. In these experiments, we set it to 60 seconds, after which the plan is considered to be invalid.

Obtained results show that taking into consideration the time limit, the proposed method outperforms the RRT* in terms of all considered criteria and gives way to SL only in terms of accuracy on the test set. Our neural-based planner generates in average the shortest paths and does it in $\SI{42}{\milli\second}$, which is at least an order of magnitude faster than other tested approaches (even excluding the running times higher than \SI{60}{\second}).
Moreover, the generation time is very stable, which is crucial in safety-critical applications.

\begin{table}[hbt]
\centering
\caption{Accuracy, execution time and length of the RRT*, SL and our path planning method on the test set (III) and accuracy of our method on training (I) and validation (II) sets. Length are reported with respect to the length of path obtained with SL. Both time and lengths statistics are reported only for valid paths.}
\begin{tabular}{c|ccc}
\hline
Planner & Accuracy [\%] & Time [s] & Length [\%]\\
\hline
RRT* & 47.82 & 10.98 $\pm$ 19.42  & 132 $\pm$ 40\\
SL & 85.63 & 0.56 $\pm$ 0.99 & 100 $\pm$ 0\\
ours I & 90.53 & - & -\\
ours II & 84.2 & - & -\\
ours III & 74.37 & 0.042$\pm$ 0.003 & 99.08 $\pm$ 3.62\\
\hline
\end{tabular}
\label{tab:results}
\end{table}

\begin{figure}[ht]
    \centering
    \includegraphics[width=\linewidth]{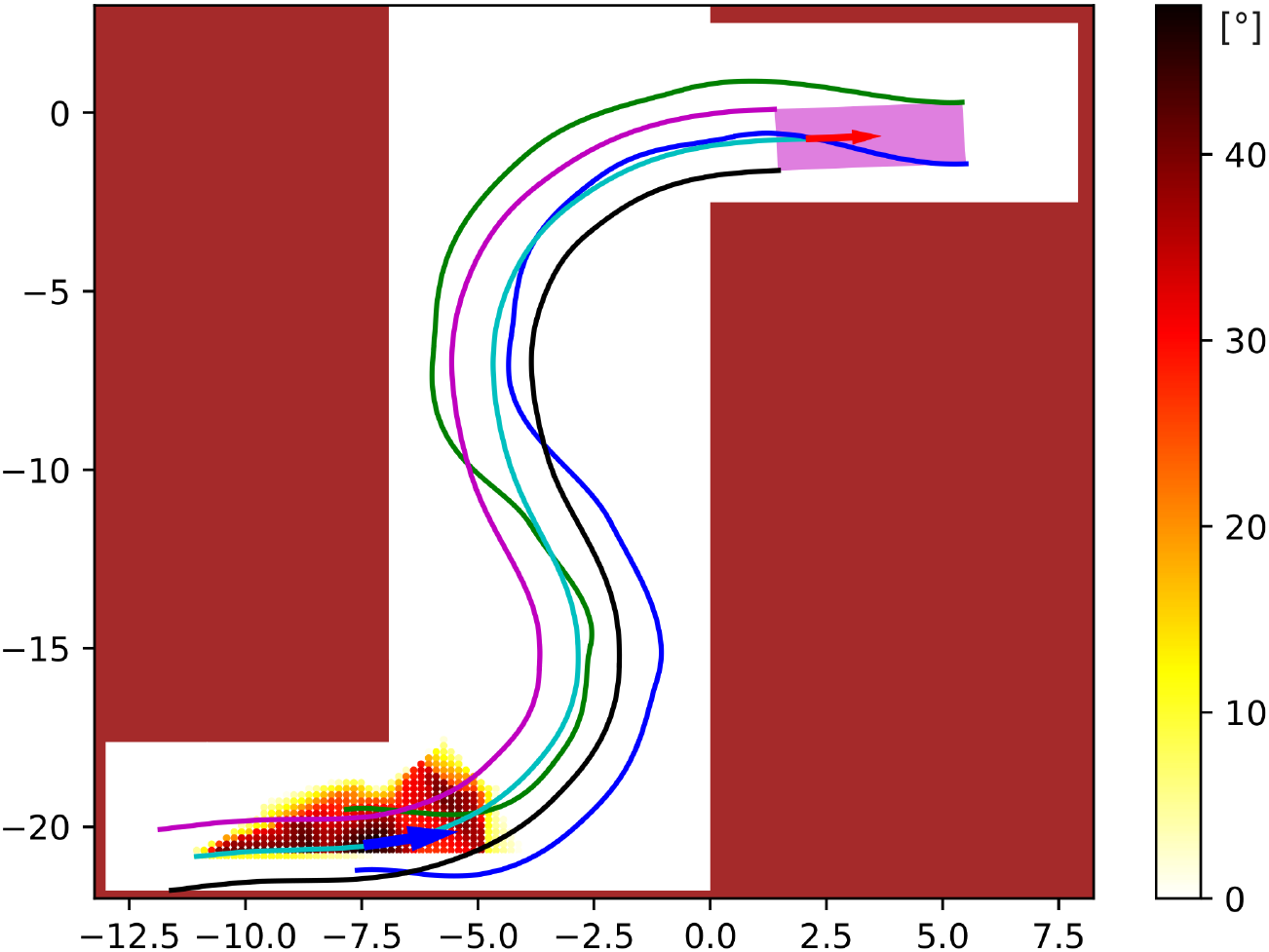}
    \caption{Visualization of the set of the initial states (denoted as the heatmap) from which it is possible to plan a feasible path to the final state (denoted with the red arrow and pink quadrangle which represents the vehicle shape). Blue arrow shows the single example of the initial state (for the given environment and goal state) which was included in the training set. Colored lines depict the paths drawn by the four corners of the vehicle and guiding point, while nominal move along the path provided by the network, for some specific initial point.}
    \label{fig:set}
\end{figure}

\begin{figure}[ht]
    \centering
    \includegraphics[width=\linewidth]{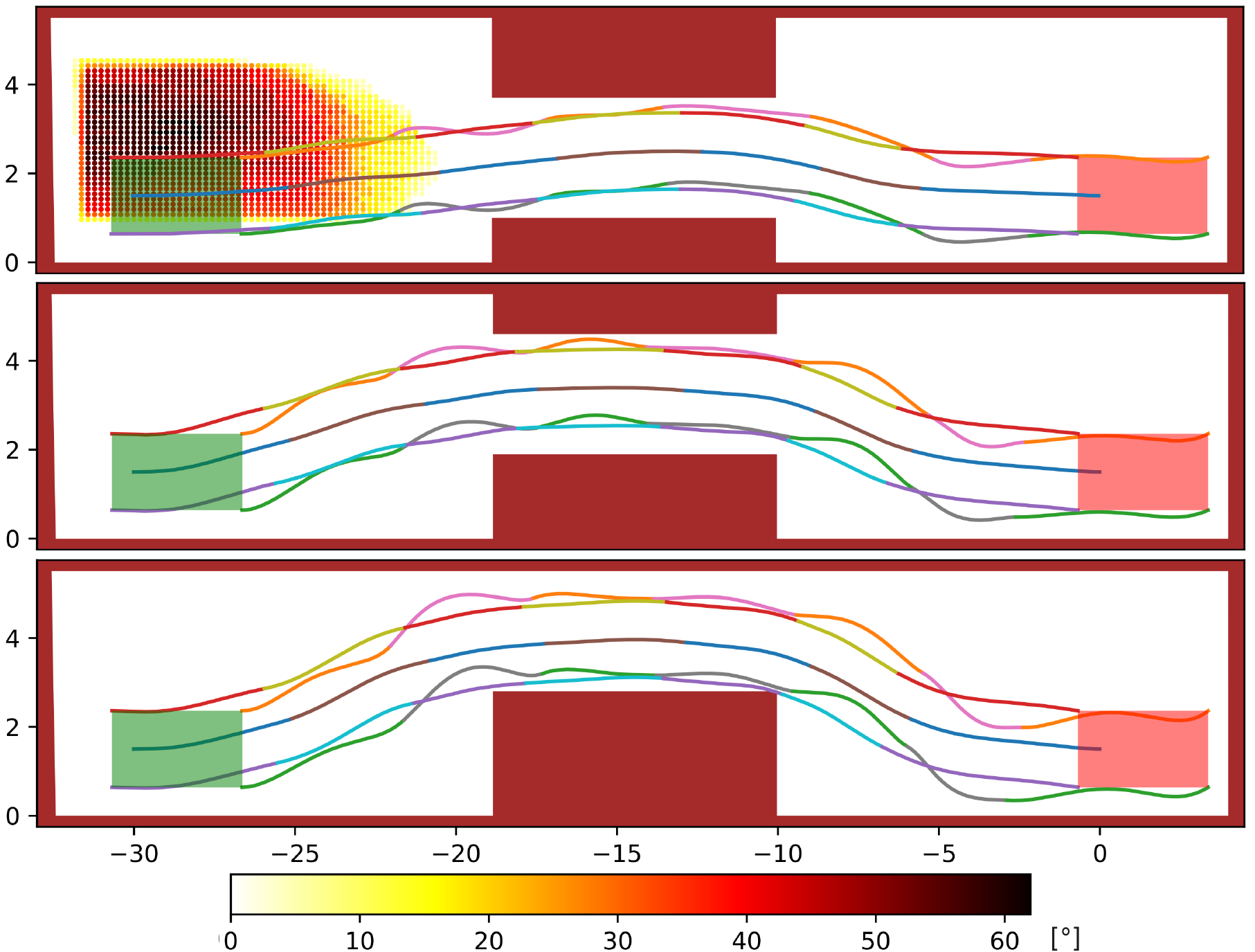}
    \caption{Visualization of the ability of the proposed neural planner to react to the changes in the environment. The above paths were obtained for the unseen scenarios. Similarly to Fig. \ref{fig:set} the colored lines depict the paths drawn by the four corners of the vehicle and guiding point, however here we show the split of the path to segments. Heatmap in the first scenario denotes the set of the initial states from which planner returned the feasible path.}
    \label{fig:geo}
    \vspace{-0.3cm}
\end{figure}

Moreover, we show the generalization abilities of the proposed approach by visualizing the set of states from which our method 
can generate feasible paths to the specific goal state.
 These sets of configurations are visualized as heatmaps for two different scenarios: perpendicular parking (Fig. \ref{fig:set}) and overtaking or parallel parking maneuver (a top scenario in Fig. \ref{fig:geo}). The environment is visualized as white and brown areas, which denotes the free and occupied areas, respectively.
In both cases, the color of the point in the heatmap indicates the range of vehicle orientations for the given point on the local map, for which the neural path planner generated feasible paths.
Using that method we obtain approximations of the configuration space size, in which our method is able to generate valid solutions.
For the environment from Figure \ref{fig:set} and the final state (denoted with the red arrow), only one example of the initial state (denoted with the blue arrow) was included in the dataset used for training. Nevertheless, the proposed neural planner is able to generalize to the much bigger set of initial states, which is depicted here as the heatmap. 

A similar situation is depicted in Figure \ref{fig:geo}, but for a completely new environment and states, which were not included in any of the datasets. It shows the robustness of the planner to the unseen tasks (initial state, final state, and environment was not included in the training set) and the ability to adjust to the modifications of the environment. This observation confirms that the network is able to reason about the geometry of the environment and adapt the generated path to the changes.

\section{Conclusions}
In this paper, we proposed a novel approach to local path planning for vehicles with kinematic constraints, which utilizes the neural network, trained in a self-supervised manner, to plan feasible paths. 
Our method contests the typical approach to motion planning based on state-space search algorithms and proposes to approximate an \textit{oracle planning function} instead. Such approach lets one to omit the time-consuming execution of the algorithm, by learning how to plan \textit{off-line} (optimizing the approximator of the \textit{oracle planning function}) and then only reusing the gathered knowledge \textit{on-line} (using a fast approximator). Although this approach gives no guarantees about the completeness, it is fast and flexible, as its quality depends on the data used in the training process and the expressiveness and generalization abilities of the model.
In our approach we do not model the \textit{oracle planning function} with the use of human-generated data, but instead, we define it implicitly by the construction of the loss function.

We examined the proposed solution using few most typical urban traffic scenarios, such as: overtaking maneuver, perpendicular and angle parking. Those tasks consist of sub-maneuvers such as passing the obstacle or parallel parking (in the overtaking task) or cornering at different angles (covered in angle and perpendicular parking), which can be combined to plan more complicated behaviors in the longer horizon.
The proposed solution achieves 74\% accuracy on the test set, which contains path planning tasks in previously unseen environments with different initial and final states. Moreover, we reported the mean planning time of the proposed method, which is equal to $\SI{42}{\milli\second}$ and compared it with popular solutions to path planning for kinematically constrained vehicles: RRT* and State Lattices. Proposed solution enables to plan at least 14 times faster than state-of-the-art methods, which opens wide range of applications in tasks that require very fast planing, but also in more general planning problems, as our method can serve as a generator of the initial solutions, which can be then used as input to other, complete planning algorithms (e.g. \cite{GENLS}. Additionally, the running times of the proposed method are very stable and the returned solution is deterministic, which is important in safety-critical systems.

Furthermore, we have presented strong generalization abilities of the proposed planner and its robustness to the changes in the environment on the selected unseen scenarios. Those experiments showed, that even the dataset contains only 15432 samples (sparse points in the task space), it is able to generalize to a relatively big neighborhood of those points. 
This result may lead to the development of some guarantees about the subsets of the task space from which the neural network planner is able to plan a valid path, which is an interesting direction of future work. Moreover, one can consider using some more flexible architectures, which will allow a variable number of segments in the path.


\addtolength{\textheight}{-12cm}   




\section*{ACKNOWLEDGMENT}
This research was partially supported by the National Center for Research and Development under the grant POIR.04.01.02-00-0081/17. Work of P. Kicki was funded from the 04/45/SBAD/0210 grant.


\bibliographystyle{IEEEtran}
\bibliography{IEEEexample}


\end{document}